\documentclass[11pt,a4paper]{article}
\usepackage[hyperref]{acl2020}
\usepackage{times}
\usepackage{latexsym}
\usepackage{bm}

\usepackage{microtype}
\usepackage{amsmath}
\usepackage{graphicx}

\aclfinalcopy 

\title{Iterative Edit-Based Unsupervised Sentence Simplification}

\author{Dhruv Kumar,$^1$ Lili Mou,$^2$ Lukasz Golab,$^1$ Olga Vechtomova$^1$\\
  $^1$University of Waterloo \\
  $^2$Department of Computing Science, University of Alberta\\ Alberta Machine Intelligence Institute (Amii)\\
  \texttt{\{d35kumar,lgolab,ovechtomova\}@uwaterloo.ca} \\
  \texttt{doublepower.mou@gmail.com}}

\date{}

\begin{document}
\maketitle
\begin{abstract}
We present a novel {iterative}, edit-based approach to unsupervised sentence simplification. Our model is guided by a scoring function involving fluency, simplicity, and meaning preservation. Then, we iteratively perform word and phrase-level edits on the complex sentence. Compared with previous approaches, our model does not require a parallel training set, but is more controllable and interpretable. Experiments on  Newsela and WikiLarge datasets show that our approach is nearly as effective as state-of-the-art supervised approaches.\footnote{Code is released at \url{https://github.com/ddhruvkr/Edit-Unsup-TS}}
\end{abstract}

\section{Introduction}

Sentence simplification is the task of rewriting text to make it easier to read, while preserving its main meaning and important information. 
Sentence simplification is relevant in various real-world and downstream applications.
For instance, it can benefit people with autism \cite{evans2014evaluation},
 dyslexia \cite{rello2013dyswebxia}, and  low-literacy skills \cite{watanabe2009facilita}.
It can also serve as a preprocessing step to improve parsers \cite{chandrasekar-etal-1996-motivations}
and summarization systems \cite{klebanov2004text}.

Recent efforts in sentence simplification have been influenced by the success of machine translation. In fact, the simplification task is often treated as monolingual translation, where a complex sentence is translated to a simple one. Such simplification systems are typically trained in a supervised way by either phrase-based machine translation \cite[PBMT,][]{wubben2012sentence, narayan2014hybrid, xu-etal-2016-optimizing} or neural machine translation  \cite[NMT,][]{zhang2017sentence, guo2018dynamic, kriz2019complexity}. Recently, sequence-to-sequence (Seq2Seq)-based NMT systems are
shown to be more successful and serve as the state of the art. 

However, supervised Seq2Seq models have two shortcomings. First, they give little insight into the simplification operations, and provide little control or adaptability to different aspects of simplification (e.g., lexical vs.~syntactical simplification).
Second, they require a large number of complex-simple aligned sentence pairs, which in turn require considerable human effort to obtain.

In previous work, researchers have addressed some of the above issues. For example, \newcite{alva2017learning} and \newcite{dong2019editnts} explicitly model simplification operators such as word insertion and deletion. Although these approaches are more controllable and interpretable than standard Seq2Seq models, they still require large volumes of aligned data to learn these operations.
To deal with the second issue, \newcite{surya2019unsupervised} recently proposed an unsupervised neural text simplification approach based on the paradigm of style transfer. 
However, their model is hard to interpret and control, like other neural network-based models. \newcite{narayan2016unsupervised} attempted to address both issues using a pipeline of lexical substitution, sentence splitting, and word/phrase deletion. However, these operations can only be executed in a fixed order. 

In this paper, we propose an iterative, edit-based unsupervised sentence simplification approach, motivated by the shortcomings of existing work.
We first design a {scoring function} that measures the quality of a candidate sentence based on the key characteristics of the simplification task, namely, fluency, simplicity, and meaning preservation. 
Then, we generate simplified candidate sentences by iteratively editing the given complex sentence using three simplification operations (lexical simplification, phrase extraction, deletion and reordering). Our model seeks the best simplified candidate sentence according to the scoring function. Compared with \newcite{narayan2016unsupervised}, the order of our simplification operations is not fixed and is decided by the model.

\begin{figure}
  \includegraphics[width=\linewidth]{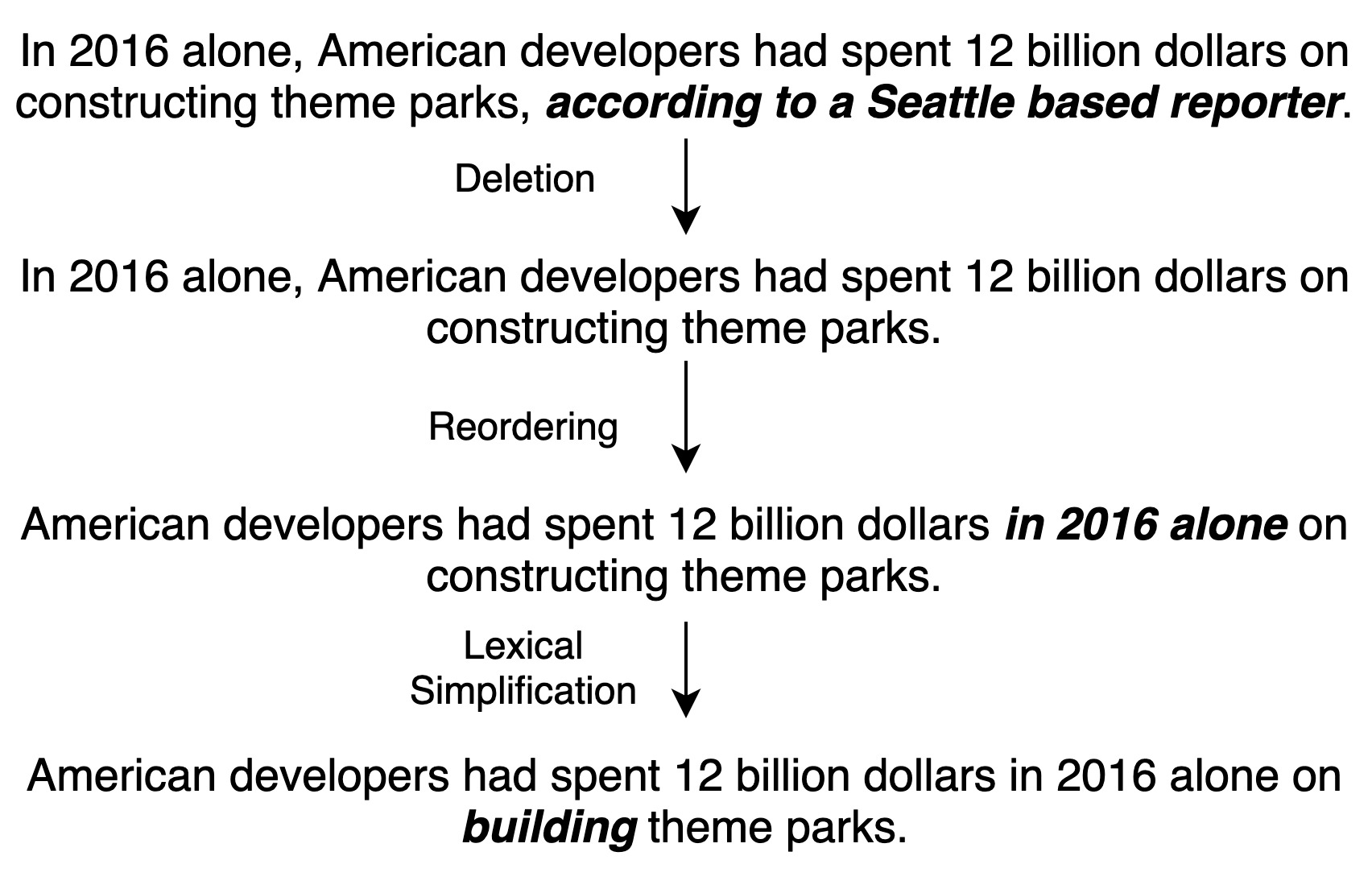}\vspace{-.3cm}
  \caption{An example of three edit operations on a given sentence. Note that dropping clauses or phrases is common in text simplification datasets.}\vspace{-.2cm}
  \label{fig:example_sentence}
\end{figure}

Figure \ref{fig:example_sentence} illustrates an example in which our model first chooses to delete a sentence fragment, followed by reordering the remaining fragments and replacing a word with a simpler synonym.

We evaluate our approach on the Newsela 
\cite{xu2015problems} and WikiLarge \cite{zhang2017sentence} corpora. 
Experiments show that our approach outperforms previous unsupervised methods and even performs competitively with state-of-the-art supervised ones, in both automatic metrics and human evaluations. We also demonstrate the interpretability and controllability of our approach, even without parallel training data.

\section{Related Work}


Early work used handcrafted rules for text simplification, at both the syntactic level~\cite{siddharthan2002architecture} and the lexicon level~\cite{carroll1999simplifying}. Later, researchers adopted machine learning methods for text simplification, modeling it as monolingual phrase-based machine translation \cite{wubben2012sentence, xu-etal-2016-optimizing}. Further, syntactic information was also considered in the PBMT framework, for example, constituency trees \cite{zhu2010monolingual} and dependency trees \cite{bingel2016text}. 
\newcite{narayan2014hybrid} performed probabilistic sentence splitting and deletion, followed by MT-based paraphrasing.



\newcite{nisioi2017exploring} employed neural machine translation (NMT) for text simplification, using a sequence-to-sequence (Seq2Seq) model \cite{sutskever2014sequence}. \newcite{zhang2017sentence} used reinforcement learning to optimize a reward based on simplicity, fluency, and relevance.
\newcite{zhao2018integrating} integrated the transformer architecture and paraphrasing rules to guide simplification learning.
\newcite{kriz2019complexity} produced diverse simplifications by generating and re-ranking candidates by fluency, adequacy, and simplicity. \newcite{guo2018dynamic} showed that simplification benefits from multi-task learning with paraphrase and entailment generation. \newcite{martin2019controllable} enhanced the transformer architecture with conditioning parameters such as length, lexical and syntactic complexity.

Recently, edit-based techniques have been developed for text simplification. \newcite{alva2017learning} trained a model to predict three simplification operators (keep, replace, and delete) from aligned pairs. 
\newcite{dong2019editnts} employed a similar approach but in an end-to-end trainable manner with neural networks. 
However, these approaches are supervised and require large volumes of parallel training data; also, their edits are only at the word level. By contrast, our method works at both word and phrase levels in an unsupervised manner. 

For unsupervised sentence simplification, \newcite{surya2019unsupervised} adopted style-transfer techniques, using adversarial and denoising auxiliary losses for content reduction and lexical simplification. However, their model is based on a Seq2Seq network, which is less interpretable and controllable. They cannot perform syntactic simplification since syntax typically does not change in style-transfer tasks. 
\newcite{narayan2016unsupervised} built a pipeline-based unsupervised framework with lexical simplification, sentence splitting, and phrase deletion. However, these operations are separate components in the pipeline, and can only be executed in a fixed order.

Unsupervised edit-based approaches have recently been explored for natural language generation tasks, such as style transfer, paraphrasing, and sentence error correction.
\newcite{li2018delete} proposed edit-based  style transfer without parallel supervision. They replaced style-specific phrases with those in the target style, which are retrieved from the training corpus. 
\newcite{miao2019cgmh} used Metropolis--Hastings sampling for constrained sentence generation. In this paper, we model text generation as a search algorithm, and design search objective and search actions specifically for text simplification. Concurrent work further shows the success of search-based unsupervised text generation for paraphrasing~\cite{UPSA} and summarization~\cite{HC}.

\section{Model}

In this section, we first provide an overview of our approach, followed by a detailed description of each component, namely, the scoring function, the edit operations, and the stopping criteria. 

\subsection{Overview}

We first define a scoring function as our search objective. It allows us to impose both hard and soft constraints, balancing the fluency, simplicity, and adequacy of candidate simplified sentences  (Section~\ref{ss:objective}).

Our approach iteratively generates multiple candidate sentences by performing a sequence of lexical and syntactic operations. It starts from the input sentence; in each iteration, it performs phrase and word edits to generate simplified candidate sentences (Section~\ref{subsec:candidates}). 

Then, a candidate sentence is selected according to certain criteria. This process is repeated until none of the candidates improve the score of the source sentence by a threshold value. The last candidate is returned as the simplified sentence (Section~\ref{sec:algorithm}).

\subsection{Scoring Function}
\label{ss:objective}

Our scoring function is the product of several individual scores that evaluate various aspects of a candidate simplified sentence. This is also known as the product-of-experts model~\cite{hinton2002training}.


\textbf{SLOR score from a syntax-aware language model (}$f_{\operatorname{eslor}}$\textbf{).}
This measures the {language fluency} and {structural simplicity} of a candidate sentence.

A probabilistic language model (LM) is often used as an estimate of sentence fluency~\cite{miao2019cgmh}.
In our work, we make two important modifications to a plain LM.

First, we replace an LM's estimated sentence probability with the syntactic log-odds ratio \cite[SLOR,][]{pauls2012large}, to better measure fluency and human acceptability.
According to \newcite{lau2017grammaticality}, SLOR shows the best correlation to human acceptability of a sentence, among many sentence probability-based scoring functions. SLOR was also shown to be effective in unsupervised text compression~\cite{kann2018sentence}.

Given a trained language model (LM) and a sentence $s$, SLOR is defined as 
\begin{equation}
\begin{split}
        \mathrm{SLOR}(s) = \frac{1}{|s|}(\mathrm{ln}(P_\mathrm{LM}(s)) - \mathrm{ln}(P_\mathrm{U}(s)))
\end{split}
\end{equation}
where $P_{\operatorname{LM}}$ is the sentence probability given by the language model, $P_{\operatorname{U}}(s)=\prod_{w\in s}P(w)$ is the product of the unigram probability of a word $w$ in the sentence, and $|s|$ is the sentence length.

SLOR essentially penalizes a plain LM's probability by unigram likelihood and the length. 
It ensures that the fluency score of a sentence is not penalized by the presence of rare words. Consider two sentences, ``\textit{I went to England for vacation}'' and ``\textit{I went to Senegal for vacation}.'' Even though both sentences are equally fluent, a standard LM will give a higher score to the former, since the word ``England'' is more likely to occur than ``Senegal.''
In simplification, SLOR is preferred for preserving rare words such as named entities.\footnote{Note that we do not use SLOR to evaluate lexicon simplicity, which will later be evaluated by the Flesch reading ease (FRE) score. The SLOR score, in fact, preserves rare words, so that we can better design dictionary-based word substitution for lexical simplification (Section~\ref{subsec:candidates}).}

Second, we use a syntax-aware LM, i.e., in addition to words, we use part-of-speech (POS) and dependency tags as inputs to the LM \cite{zhao2018language}. For a word $w_i$, the input to the syntax-aware LM is $[\bm e(w_i);\bm p(w_i);\bm d(w_i)]$, where $\bm e(w_i)$ is the word embedding, $\bm p(w_i)$ is the POS tag embedding, and $\bm d(w_i)$ is the dependency tag embedding.

Note that our LM is trained on simple sentences. Thus, the syntax-aware LM prefers a syntactically simple sentence. It also helps to identify sentences that are structurally ungrammatical. 

\textbf{Cosine Similarity (}$f_{\cos}$\textbf{).} Cosine similarity is an important measure of meaning preservation. We compute the cosine value between sentence embeddings of the original complex sentence ($c$) and the generated candidate sentence ($s$), where our sentence embeddings are calculated as the idf weighted average of individual word embeddings. Our sentence similarity measure acts as a hard filter, i.e., $f_\text{cos}(s)=1$ if $\cos(\bm c,\bm s)>\tau$, or $f_\text{cos}(s)=0$ otherwise, for some threshold~$\tau$.

\textbf{Entity Score (}$f_{\operatorname{entity}}$\textbf{).} Entities help identify the key information of a sentence and therefore are also useful in measuring {meaning preservation}. Thus, we count the number of entities in the sentence as part of the scoring function, where entities are detected by a third-party tagger.

\textbf{Length (}$f_{\operatorname{len}}$\textbf{).}  This score is proportional to the inverse of the sentence length. It forces the model to generate {shorter} and {simpler} sentences. However, we reject sentences shorter than a specified length ($\le$6 tokens) to prevent over-shortening.

\textbf{FRE (}$f_{\operatorname{fre}}$\textbf{).} The Flesch Reading Ease (FRE) score \cite{kincaid1975derivation} measures the ease of readability in text. It is based on text features such as the average sentence length and the average number of syllables per word. A higher scores indicate that the text is \textit{simpler} to read.

\smallskip
We compute the overall scoring function as the product of individual scores.
\begin{equation}
\begin{split}
        {f}(s) = f_\mathrm{{eslor}}(s)^\alpha \cdot f_\mathrm{{fre}}(s)^\beta \cdot (1/f_\mathrm{{len}}(s))^\gamma \\ 
         \cdot f_\mathrm{{entity}}(s)^\delta \cdot f_\mathrm{{cos}}(s)  \\
\end{split}
\end{equation}
where the weights $\alpha$, $\beta$, $\gamma$, and $\delta$ balance the relative importance of the different scores. Recall that the cosine similarity measure does not require a weight since it is a hard indicator function.

In Section \ref{adaptability}, we will experimentally show that the weights defined for different scores affect different characteristics of simplification and thus provide more adaptability and controllability.

\subsection{Generating Candidate Sentences}
\label{subsec:candidates}

\begin{figure}\centering
  \includegraphics[width=.9\linewidth]{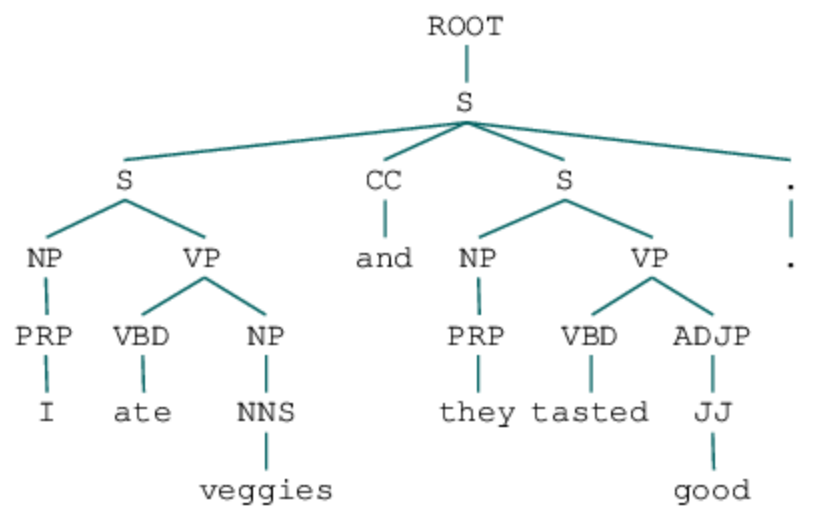}\vspace{-.2cm}
  \caption{Constituency parse tree is used for detecting phrases.}\vspace{-.2cm}
  \label{fig:example_parsetree}
\end{figure}

We generate candidate sentences by editing words and phrases. We use a third-party parser to obtain the constituency tree of a source sentence. Each clause- and phrase-level constituent (e.g., S, VP, and NP) is considered as a phrase. 
Since a constituent can occur at any depth in the parse tree, we can deal with both long and short phrases at different granularities. 
In Figure \ref{fig:example_parsetree}, for example, both  ``\textit{good}'' (ADJP) and ``\textit{tasted good}'' (VP) are constituents and thus considered as phrases, whereas ``\textit{tasted}'' is considered as a single word. 
For each phrase, we generate a candidate sentence using the edit operations explained below, with Figure \ref{fig:example_sentence} being a running example.


\textbf{Removal.}
For each phrase detected by the parser, this operation generates a new candidate sentence by removing that phrase from the source sentence. In Figure~\ref{fig:example_sentence}, our algorithm can drop the phrase
``\textit{according to a Seattle based reporter},'' which is not the main clause of the sentence. The removal operation allows us to remove peripheral information in a sentence for content reduction. 
 
\textbf{Extraction.}
This operation simply extracts a selected phrase (including a clause) as the candidate sentence. This allows us to select the main clause in a sentence and remove remaining peripheral information.

\textbf{Reordering.}
For each phrase in a sentence, we generate candidate sentences by moving the phrase before or after another phrase (identified by clause- and phrase-level constituent tags).
In the running example, the phrase ``\textit{In 2016 alone}'' is moved between the phrases ``\textit{12 billion dollars}'' 
and ``\textit{on constructing theme parks}.''
As seen, the reordering operation is able to perform syntactic simplification.

\textbf{Substitution.}
In each phrase, we identify the most complex word as the rarest one according to the idf score. 
For the selected complex word, we generate possible substitutes using a two-step strategy.

First, we obtain candidate synonyms by taking the union of the WordNet synonym set \cite{miller1995wordnet} and the closest words from GloVe~\cite{pennington2014glove} and Word2Vec~\cite{mikolov2013distributed} embeddings (where embedding closeness is measured by Euclidean distance).  Second, a candidate synonym is determined to be an appropriate simple substitute if it satisfies the following conditions: a) it has a lower idf score than the complex word, where the scores are computed from the target simple sentences, 
b) it is not a morphological inflection of the complex word, c) its word embedding exceeds a cosine similarity  threshold to the complex word, 
and, d) it is has the same part-of-speech and dependency tags in the sentence as the complex word. We then generate candidate sentences by replacing the complex word with all qualified lexical substitutes.
Notably, we do not replace entity words identified by entity taggers. 

In our example sentence, consider the phrase ``\textit{constructing theme parks}.'' 
The word ``constructing'' is chosen as the word to be simplified, and is replaced with ``building.'' As seen, this operation performs lexical simplification.

\subsection{The Iterative Algorithm}
\label{sec:algorithm}

Given an input complex sentence, our algorithm iteratively performs edits to search for a higher-scoring candidate. 

In each iteration, we consider all the operations (i.e., removal, extraction, reordering, and substitution). Each operation may generate multiple candidates (e.g., multiple words for substitution); we filter out a candidate sentence if the improvement does not pass an operation-specific threshold. We choose the highest-scoring sentence from those that are not filtered out. Our algorithm terminates if no edit passes the threshold, and the final candidate is our generated simplified sentence.

Our algorithm includes a filtering step for each operation. We only keep a candidate sentence if it is better than the previous one by a multiplicative factor, i.e., 
\begin{equation}\label{eqn:stop}
{f}(c)/{f}(s) > r_{\operatorname{op}}
\end{equation} where $s$ is the sentence given by the previous iteration, and $c$ is a candidate generated by operator ${\operatorname{op}}$ from $s$.

Notably, we allow different thresholds for each operation.  This provides control over different aspects of simplification, namely, lexicon simplification, syntactic simplification, and content reduction. A lower threshold for substitution, for example, encourages the model to perform more lexical simplification. 

\section{Experiments}

\subsection{Data}
\label{ssec:layout}

We use the Newsela \cite{xu2015problems} and the WikiLarge datasets \cite{zhang2017sentence} for evaluating our model. 

Newsela is a collection of 1,840 news articles written by professional editors at 5 reading levels for children.  We use the standard split and exclude simple-complex sentence pairs that are one reading level apart, following \newcite{zhang2017sentence}. 
This gives 95,208 training, 1,129 validation, and 1,077 test sentences. 

The WikiLarge dataset is currently the largest text simplification corpus. It contains 296,402, 2,000, and 359 complex-simple sentence pairs for training, validation, and testing, respectively. The training set of WikiLarge consists of automatically aligned sentence pairs from the normal and simple Wikipedia versions. The validation and test sets contain multiple human-written references, against which we evaluate our algorithm. 

For each corpus, we only use its training set to learn a language model of simplified sentences. {For the WikiLarge dataset,} we also train a Word2Vec embedding model from scratch on its source and target training sentences. These embeddings are used to obtain candidate synonyms in the substitution operation. 

\subsection{Training Details}

For the LM, we use a two-layer, 256-dimensional recurrent neural network (RNN) with the gated recurrent unit %
\cite[GRU,][]{chung2014empirical}.
We initialize word embeddings using 300-dimensional GloVe \cite{pennington2014glove}; out-of-vocabulary words are treated as \texttt{UNK}, initialized uniformly in the range of $\pm 0.05$. Embeddings for POS tags and dependency tags are 150-dimensional, also initialized randomly. We fine-tune all embeddings during training.

We use the Averaged Stochastic Gradient Descent (ASGD) algorithm \cite{polyak1992acceleration} to train the LM, with $0.4$ as the dropout and $32$ as the batch size. For the Newsela dataset, the thresholds $r_{\operatorname{op}}$ in the scoring function are set to $1.25$ for all the edit operations. All the weights in our scoring function ($\alpha, \beta,\gamma,\delta$) are set to $1$. For the WikiLarge dataset, the thresholds are set as $1.25$ for the removal and reordering operations, $0.8$ for substitution, and $5.0$ for extraction. The weights in the scoring function ($\alpha,\beta,\gamma,\delta$) are set to $0.5, 1.0, 0.25$ and $1.0$, respectively.  

We use CoreNLP \cite{manning2014stanford} to construct the constituency tree and Spacy\footnote{\url{https://spacy.io/}} to generate part-of-speech and dependency tags.

\begin{table*}[t!]
\begin{center}
\resizebox{.85\textwidth}{!}{
\begin{tabular}{|l|c|c|c|c|c|c|c|c|}
\hline \textbf{Method} & \textbf{SARI}$^\uparrow$ & \textbf{Add}$^\uparrow$ & \textbf{Delete}$^\uparrow$ &\textbf{Keep}$^\uparrow$ & \textbf{BLEU}$^\uparrow$ & \textbf{GM}$^\uparrow$ & \textbf{FKGL}$^\downarrow$ & \textbf{Len}  \\ \hline
Reference & 70.13 & - & - & - & 100 & 83.74 & 3.20 & 12.75 \\
\hline
 \multicolumn{9}{|c|}{Baselines} \\
\hline
Complex & 2.82 & - & - & - & 21.30 & 7.75  & 8.62 & 23.06 \\
Reduce-250 & 28.39 & - & - & - & 11.79 & 18.29 & -0.23 & 14.48 \\
\hline
 \multicolumn{9}{|c|}{Supervised Methods} \\
\hline
PBMT-R & 15.77 & 3.07 & 38.34 & 5.90 & 18.1 & 16.89 & 7.59 & 23.06 \\
Hybrid & 28.61\text{*} & 0.95\text{*} & 78.86\text{*} & 6.01\text{*} & 14.46 & 20.34 & 4.03 & 12.41 \\
EncDecA & 24.12 & 2.73 & 62.66 & 6.98 & 21.68 & 22.87 & 5.11 & 16.96 \\
Dress & 27.37 & \textbf{3.08} & 71.61 & 7.43 & 23.2 & 25.2 & 4.11 & 14.2 \\
Dress-Ls & 26.63 & 3.21 & 69.28 & 7.4 & \textbf{24.25} & 25.41 & 4.21 & 14.37 \\
DMass & 31.06 & 1.25 & 84.12 & 7.82 & 11.92 & 19.24 & 3.60 & 15.07 \\
S2S-All-FA & 30.73 & 2.64 & 81.6 & \textbf{7.97} & 19.55 & 24.51 & 2.60 & 10.81 \\
Edit-NTS & 30.27\text{*} & 2.71\text{*} & 80.34\text{*} & 7.76\text{*} & 19.85 & 24.51 & 3.41 & 10.92 \\
EncDecP & 28.31 & - & - & - & 23.72 & \textbf{25.91} & - & - \\
EntPar & \textbf{33.22} & 2.42 & \textbf{89.32} & 7.92 & 11.14 & 19.24 & 1.34 & 7.88 \\
\hline
 \multicolumn{9}{|c|}{Unsupervised Methods (Ours)} \\
\hline
RM+EX & 26.07 & 2.35 & 68.35 & 7.5 & \textbf{27.22} & 26.64 & 2.95 & 12.9 \\
RM+EX+LS & 26.26 & 2.28 & 68.94 & 7.57 & 27.17 & \textbf{26.71} & 2.93 & 12.88 \\
RM+EX+RO & 26.99 & \textbf{2.47} & 70.88 & 7.63 & 26.31 & 26.64 & 3.14 & 12.81 \\
RM+EX+LS+RO & 27.11 & 2.40 & 71.26 & \textbf{7.67} & 26.21 & 26.66 & 3.12 & 12.81 \\
RM+EX+LS+RO$^{\dagger}$ & \textbf{30.44} & 2.05 & \textbf{81.77} & 7.49 & 17.36 & 22.99 & 2.24 & 9.61 \\
\hline
\end{tabular}
}
\end{center}\vspace{-.3cm}
\caption{\label{results-Newsela} Results on the Newsela dataset. $^{\dagger}$ denotes the model with parameters tuned by SARI; other variants are tuned by the geometric mean (GM). $^\uparrow$The higher, the better. $^\downarrow$The lower, the better. * indicates a number that is different from that reported in the original paper. This is due to a mistreatment of capitalization in the previous work (confirmed by personal correspondence).}\vspace{-.2cm}
\end{table*}

\subsection{Competing Methods}

We first consider the reference to obtain an upper-bound for a given evaluation metric. 
We also consider the complex sentence itself as a trivial baseline, denoted by \texttt{Complex}.  

Next, we develop a simple heuristic that removes rare words occurring $\le$ 250 times in the simple sentences of the training corpus, denoted by \texttt{Reduce-250}. As discussed in Section~\ref{ss:results}, this simple heuristic demonstrates the importance of balancing different automatic evaluation metrics. 

For unsupervised competing methods, we compare with \newcite{surya2019unsupervised}, which is inspired by unsupervised neural machine translation. They proposed two variants, \texttt{UNMT} and \texttt{UNTS}, but their results are only available for WikiLarge.

We also compare our model with supervised methods.
First, we consider non-neural phrase-based machine translation (PBMT) methods: {\tt PBMT-R} \cite{wubben2012sentence}, which re-ranks sentences generated by PBMT for diverse simplifications; {\tt SBMT-SARI} \cite{xu-etal-2016-optimizing}, which uses an external paraphrasing database; and {\tt Hybrid} \cite{narayan2014hybrid}, which uses a combination of PBMT and discourse representation structures.
Next, we compare our method with neural machine translation (NMT) systems: {\tt EncDecA}, which is a vanilla Seq2Seq model with attention \cite{nisioi2017exploring}; {\tt Dress} and {\tt Dress-Ls}, which are based on deep reinforcement learning \cite{zhang2017sentence}; {\tt DMass}  \cite{zhao2018integrating}, which is a transformer-based model with external simplification rules; {\tt EncDecP}, which is an encoder-decoder model with a pointer-mechanism; {\tt EntPar}, which is based on multi-task learning \cite{guo2018dynamic}; {\tt S2S-All-FA}, which a reranking based model focussing on lexical simplification \cite{kriz2019complexity}; and {\tt Access}, which is based on the transformer architecture \cite{martin2019controllable}. Finally, we compare with a supervised edit-based neural model, {\tt Edit-NTS} \cite{dong2019editnts}.

We evaluate our model with a different subset of operations, i.e., removal ({\tt RM}), extraction ({\tt EX}), reordering ({\tt RO}), and lexical substitution ({\tt LS}). 
In our experiments, we test the following variants: {\tt RM+EX}, {\tt RM+EX+LS}, {\tt RM+EX+RO}, and {\tt RM+EX+LS+RO}.

\begin{table*}[t!]
\begin{center}
\resizebox{.8\textwidth}{!}{
\begin{tabular}{|l|c|c|c|c|c|c|c|}
\hline \textbf{Method} & \textbf{SARI}$^\uparrow$ & \textbf{Add}$^\uparrow$ & \textbf{Delete}$^\uparrow$ &\textbf{Keep}$^\uparrow$ & \textbf{BLEU}$^\uparrow$ & \textbf{FKGL}$^\downarrow$ & \textbf{Len}  \\
\hline
 \multicolumn{8}{|c|}{Baselines} \\
\hline
Complex & 27.87 & - & - & - & 99.39 & - & 22.61 \\
\hline
 \multicolumn{8}{|c|}{Supervised Methods} \\
\hline
PBMT-R & 38.56 & 5.73 & 36.93 & 73.02 & 81.09 & 8.33 & 22.35 \\
Hybrid & 31.40 & 1.84 & 45.48 & 46.87 & 48.67 & \textbf{4.56} & 13.38 \\
EncDecA & 35.66 & 2.99 & 28.96 & \textbf{75.02} & \textbf{89.03} & 8.42 & 21.26 \\
Dress & 37.08 & 2.94 & 43.15 & 65.15 & 77.41 & 6.59 & 16.14 \\
Dress-Ls & 37.27 & 2.81 & 42.22 & 66.77 & 80.44 & 6.62 & 16.39 \\
Edit-NTS & 38.23 & 3.36 & 39.15 & 72.13 & 86.69 & 7.30 & 18.87 \\
EntPar & 37.45 & - & - & - & 81.49 & 7.41 & - \\
Access & \textbf{41.87} & \textbf{7.28} & \textbf{45.79} & 72.53 & 75.46 & 7.22 & 22.27 \\
\hline
 \multicolumn{8}{|c|}{Models using external knowledge base} \\
\hline
SBMT-SARI & 39.96 & 5.96 & 41.42 & 72.52 & 73.03 & 7.29 & 23.44 \\
DMass & 40.45 & 5.72 & 42.23 & 73.41 & - & 7.79 & - \\
\hline
 \multicolumn{8}{|c|}{Unsupervised Methods} \\
\hline
UNMT & 35.89 & 1.94 & 37.68 & 68.04 & 70.61 & 8.23 & 21.85 \\
UNTS & 37.20 & 1.50 & 41.27 & 68.81 & 74.02 & 7.84 & 19.05 \\
\hline
RM+EX & 36.46 & 1.68 & 35.17 & \textbf{72.54} & \textbf{88.90} & 6.47 & 18.62 \\
RM+EX+LS & \textbf{37.85} & \textbf{2.31} & 43.65 & 67.59 & 73.62 & \textbf{6.30} & 18.45 \\
RM+EX+RO & 36.54 & 1.73 & 36.10 & 71.79 & 85.07 & 6.89 & 19.24 \\
RM+EX+LS+RO & 37.58 & 2.30 & \textbf{43.97} & 66.46 & 70.15 & 6.69 & 19.54 \\
\hline
\end{tabular}}
\end{center}\vspace{-.4cm}
\caption{\label{results-Wikilarge} Results on the WikiLarge dataset. $^\uparrow$The higher, the better. $^\downarrow$The lower, the better. }\vspace{-.1cm}
\end{table*}

\subsection{Automatic Evaluation}
\label{ss:results}

Tables \ref{results-Newsela} and \ref{results-Wikilarge} present the results of the automatic evaluation on the Newsela and WikiLarge datasets, respectively.

We use the SARI metric \cite{xu-etal-2016-optimizing} to measure the simplicity of the generated sentences. SARI computes the arithmetic mean of the $n$-gram F1 scores of three rewrite operations: adding, deleting, and keeping. The individual F1-scores of these operations are reported in the columns ``Add,'' ``Delete,'' and ``Keep.''

We also compute the BLEU score \cite{papineni2002bleu} to measure the closeness between a candidate and a reference. \newcite{xu-etal-2016-optimizing} and \newcite{sulem2018bleu} show that BLEU correlates with human judgement on fluency and meaning preservation for text simplification.\footnote{This does not hold when sentence splitting is involved. In our datasets, however, sentence splitting is rare, for example, 0.18\% in the Newsela validation set).}

In addition, we include a few intrinsic measures (without reference) to evaluate the quality of a candidate sentence: the Flesch--Kincaid grade level (FKGL) evaluating the ease of reading, as well as the average length of the sentence.

A few recent text simplification studies~\cite{dong2019editnts, kriz2019complexity} did not use BLEU for evaluation, noticing that the complex sentence itself achieves a high BLEU score (albeit a low SARI score), since the complex sentence is indeed fluent and preserves meaning. This is also shown by our {\tt Complex} baseline.

For the Newsela dataset, however, we notice that the major contribution to the SARI score is from the deletion operation. By analyzing previous work such as {\tt EntPar}, we find that it reduces the sentence length to a large extent, and achieves high SARI due to the extremely high F1 score of ``Delete.'' However, its BLEU score is low, showing the lack of fluency and meaning. This is also seen from the high SARI of ({\tt Reduce-250}) in Table \ref{results-Newsela}. 
Ideally, we want both high SARI and high BLEU, and thus, we calculate the geometric mean (GM) of them as the main evaluation metric for the Newsela dataset. 
\begin{table*}[t!]
\begin{center}
\resizebox{.85\linewidth}{!}{
\begin{tabular}{|l|c|c|c|c|c|c|c|c|}
\hline \textbf{Method} & \textbf{SARI}$^\uparrow$ & \textbf{Add}$^\uparrow$ & \textbf{Delete}$^\uparrow$ &\textbf{Keep}$^\uparrow$ & \textbf{BLEU}$^\uparrow$ & \textbf{GM}$^\uparrow$ & \textbf{FKGL}$^\downarrow$ & \textbf{Len} \\ \hline
RM+EX+LS+RO & 27.11 & 2.40 & 71.26 & 7.67 & 26.21 & 26.66 & 3.12 & 12.81 \\
\hline
$-$ SLOR & 27.63 & 2.22 & 73.20 & 7.49 & 24.14 & 25.83 & 2.61 & 12.37 \\
$-$ syntax-awareness & 26.91 & 2.16 & 71.19 & 7.39 & 24.98 & 25.93 & 3.65 & 12.76 \\
\hline
\end{tabular}}
\end{center}\vspace{-.4cm}
\caption{\label{results-designchoices} Ablation test of the SLOR score based on syntax-aware language modeling.}
\end{table*}

\begin{table}[t!]
\begin{center}
\resizebox{.9\linewidth}{!}{
\begin{tabular}{|l|c|c|c|c|c|}
\hline \textbf{Value} & \textbf{SARI}$^\uparrow$ & \textbf{BLEU}$^\uparrow$ & \textbf{GM}$^\uparrow$  & \textbf{FRE}$^\uparrow$ & \textbf{Len}  \\
\hline
 \multicolumn{6}{|c|}{Effect of threshold $r_{\operatorname{op}}$} \\
\hline
1.0 & 29.20 & 21.69 & 25.17 & 83.75 & 11.75 \\
1.1 & 28.38 & 23.59 & 25.87 & 82.83 & 12.17 \\
1.2 & 27.45 & 25.54 & 26.48 & 81.98 & 12.62 \\
1.3 & 26.60 & \textbf{26.47} & 26.53 & 81.47 & 13.07 \\
\hline
 \multicolumn{6}{|c|}{Effect of weight $\alpha$ for $f_{\rm eslor}$} \\
\hline
0.75 & 27.04 & 25.75 & 26.39 & 83.46 & 12.46 \\
1.25 & 26.91 & 25.96 & 26.43 & 81.26 & 12.96 \\
1.50 & 26.74 & 25.20 & 25.96 & 80.94 & 13.06 \\
2.0 & 26.83 & 24.29 & 25.53 & 80.11 & 13.15 \\
\hline
 \multicolumn{6}{|c|}{Effect of weight $\beta$ for $f_{\rm fre}$} \\
\hline
0.5 & 26.42 & 25.53 & 25.97 & 78.61 & 13.20 \\
1.5 & 27.38 & 26.04 & \textbf{26.70} & 84.31 & 12.58 \\
2.0 & 27.83 & 25.27 & 26.52 & 87.03 & 12.26 \\
3.0 & 28.29 & 23.69 & 26.52 & \textbf{90.34} & 11.91 \\
\hline
 \multicolumn{6}{|c|}{Effect of weight $\gamma$ for $1/f_{\rm len}$} \\
\hline
0.5 & 24.54 & 25.06 & 24.80 & 80.49 & 14.55 \\
2.0 & 29.00 & 21.65 & 25.06 & 82.69 & 10.93 \\
3.0 & 29.93 & 19.05 & 23.88 & 82.20 & 10.09 \\
4.0 & \textbf{30.44} & 17.36 & 22.99 & 80.86 & 9.61 \\
\hline
 \multicolumn{6}{|c|}{Effect of weight $\delta$ for $f_{\rm entity}$} \\
\hline
0.5 & 27.81 & 24.68 & 26.20 & 83.6 & 12.01 \\
2.0 & 25.44 & 24.63 & 25.03 & 79.36 & 14.28 \\
\hline
\end{tabular}}
\end{center}\vspace{-.2cm}
\caption{\label{adaptable} Analysis of the threshold value of the stopping criteria and relative weights in the scoring function. }\vspace{-.2cm}
\end{table}
On the other hand, this is not the case for WikiLarge, since none of the models can achieve high SARI by using only one operation among ``Add,'' ``Delete,'' and ``Keep.''
Moreover, the complex sentence itself yields an almost perfect BLEU score (partially due to the multi-reference nature of WikiLarge). Thus, we do not use GM, and for this dataset, SARI is our main evaluation metric. 

\textbf{Overall results on Newsela.}
 Table~\ref{results-Newsela} shows the results on Newsela. By default (without $\dag$), validation is performed using the GM score. Still, our unsupervised text simplification achieves a SARI score around 26--27, outperforming quite a few supervised methods. Further, we experiment with SARI-based validation (denoted by $\dag$), following the setting of most previous work~\cite{dong2019editnts,guo2018dynamic}. We achieve 30.44 SARI, which is competitive with state-of-the-art supervised methods.

Our model also achieves high BLEU scores. As seen, all our variants, if validated by GM (without~$\dag$), outperform competing methods in BLEU. One of the reasons is that our model performs text simplification by making edits on the original sentence instead of rewriting it from scratch.

In terms of the geometric mean (GM), our unsupervised approach outperforms all previous work, showing a good balance between simplicity and content preservation. The readability of our generated sentences is further confirmed by the intrinsic FKGL score.

\textbf{Overall results on WikiLarge.}
For the Wikilarge experiments in Table~\ref{results-Wikilarge}, we perform validation on SARI, which is the main metric in this experiment. Our model outperforms existing unsupervised methods, and is also competitive with  state-of-the-art supervised methods. 

We observe that lexical simplification ({\tt LS}) is important in this dataset, as its improvement is large compared with the Newsela experiment in Table~\ref{results-Newsela}.
 Additionally, reordering ({\tt RO}) does not improve performance, as it is known that WikiLarge does not focus on syntactic simplification~\cite{xu-etal-2016-optimizing}.
The best performance for this experiment is obtained by the {\tt RM+EX+LS} model.

\subsection{Controllability}
\label{adaptability}

We now perform a detailed analysis of the scoring function described in Section~\ref{ss:objective} to understand the effect on different aspects of simplification.  We use the  {\tt RM+EX+LS+RO} variant and the Newsela corpus as the testbed.

\textbf{The SLOR score with syntax-aware LM.}
We analyze our syntax-aware SLOR score in the search objective. First, we remove the SLOR score and use the standard sentence probability. We observe that SLOR helps preserve rare words, which may be entities. As a result,
the readability score (FKGL) becomes better (i.e., lower), but the BLEU score decreases. 
We then evaluate the importance of using a structural LM instead of a standard LM. We see a decrease in both SARI and BLEU scores. In both cases, the GM score decreases.

\textbf{Threshold values and relative weights.}
Table~\ref{adaptable} analyzes the effect of the hyperparameters of our model, namely, the threshold in the stopping criteria and the relative weights in the scoring function.

As discussed in Section~\ref{sec:algorithm}, we use a threshold as the stopping criteria for our iterative search algorithm. For each operation, we require that a new candidate should be better than the previous iteration by a multiplicative threshold $r_{\operatorname{op}}$ in Equation~(\ref{eqn:stop}). In this analysis, we set the same threshold for all operations for simplicity.
As seen in Table~\ref{adaptable}, 
increasing the threshold leads to better meaning preservation since the model is more conservative (making fewer edits). This is shown by the higher BLEU and lower SARI scores. 

Regarding the weights for each individual scoring function, we find that increasing the weight~$\beta$ for the FRE readability score makes sentences shorter, more readable, and thus simpler. This is also indicated by  higher SARI values. 
When sentences are rewarded for being short (with large $\gamma$), SARI increases but BLEU decreases, showing less meaning preservation. 
The readability scores initially increase with the reduction in length, but then decrease. 
Finally, if we increase the weight $\delta$ for the entity score, the sentences become longer and more complex since the model is penalized more for deleting entities. 

In summary, the above analysis shows the controllability of our approach in terms of different simplification aspects, such as simplicity, meaning preservation, and readability.

\subsection{Human Evaluation}

We conducted a human evaluation on the Newsela dataset  since automated metrics may be insufficient for evaluating text generation. We chose 30 sentences from the test set for annotation and considered a subset of baselines. For our model variants, we chose {\tt RM+EX+LS+RO}, considering both validation settings (GM and SARI). 

We followed the evaluation setup in \newcite{dong2019editnts}, and measure the adequacy (\textit{How much meaning from the original sentence is preserved?}), simplicity (\textit{Is the output simper than the original sentence?}), and fluency (\textit{Is the output grammatical?}) on a five-point Likert scale.  We recruited three volunteers, one native English speaker and two non-native fluent English speakers. Each of the volunteer was given 30 sentences from different models (and references) in a randomized order. Additionally, we asked the volunteers to measure the number of instances where models produce incorrect details or generate text that is not implied by the original sentence.
We did this because neural models are known to hallucinate information \cite{rohrbach2018object}.
We report the average count of false information per sentence, denoted as~FI.

We observe that our model {\tt RM+EX+LS+RO} (when validated by GM) performs better than {\tt Hybrid}, a combination of PBMT and discourse representation structures, in all aspects. It also performs competitively with remaining supervised NMT models. 

For adequacy and fluency, {\tt Dress-Ls} performs the best since it produces relatively longer sentences. For simplicity, {\tt S2S-All-FA} performs the best since it produces shorter sentences. 
Thus, a balance is needed between these three measures. As seen, {\tt RM+EX+LS+RO} ranks second in terms of the average score in the list (reference excluded). The human evaluation confirms the effectiveness of our unsupervised text simplification, even when compared with supervised methods.

We also compare our model variants {\tt RM+EX+LS+RO} (validated by GM) and {\tt RM+EX+LS+RO$^{\dagger}$} (validated by SARI). As expected, the latter generates shorter sentences, performing better in simplicity but worse in adequacy and fluency. 

Regarding false information (FI), we observe that previous neural models tend to generate more false information, possibly due to the vagueness in the continuous space. 
By contrast, our approach only uses neural networks in the scoring function, but performs discrete edits of words and phrases. Thus, we achieve high fidelity (low FI) similar to the non-neural {\tt Hybrid} model, which also performs editing on discourse parsing structures with PBMT.

In summary, our model takes advantage of both neural networks (achieving high adequacy, simplicity, and fluency) and traditional phrase-based approaches (achieving high fidelity).

Interestingly, the reference of Newsela has a poor (high) FI score, because the editors wrote simplifications at the document level, rather than the sentence level.

\begin{table}[t]
\centering
\resizebox{\linewidth}{!}{
\begin{tabular}{|c|c|c|c|c|c|}
\hline
\textbf{Method} & \textbf{A}$^\uparrow$ & \textbf{S}$^\uparrow$ & \textbf{F}$^\uparrow$ & \textbf{Avg}$^\uparrow$ & \textbf{FI}$^\downarrow$ \\
\hline
\small{Hybrid} & 2.63 & 2.74 & 2.39 & 2.59 & \textbf{0.03} \\
\small{Dress-Ls} & \textbf{3.29} & 3.05 & \textbf{4.11} & \textbf{3.48} & 0.2 \\
\small{EntPar} & 1.92 & 2.97 & 3.16 & 2.68 & 0.47 \\
\small{S2S-All-FA} & 2.25 & \textbf{3.24} & 3.90 & 3.13 & 0.3 \\
\small{Edit-NTS} & 2.37 & 3.17 & 3.73 & 3.09 & 0.23 \\
\hline
\small{RM+EX+LS+RO} & 2.97 & 3.09 & 3.78 & 3.28 & \textbf{0.03} \\
\small{RM+EX+LS+RO$^{\dagger}$} & 2.58 & 3.21 & 3.33 & 3.04 & 0.07 \\
\hline
\small{Reference} & 2.91 & 3.49	& 4.46 & 3.62 & 0.77 \\
\hline
\end{tabular}}
\caption{Human evaluation on Newsela, where we measure adequacy (A), simplicity (S), fluency (F), and their average score (Avg), based on 1--5 Likert scale. We also count average instances of false information per sentence (FI).}\vspace{-.2cm}
\end{table}

\section{Conclusion}

We proposed an iterative, edit-based approach to text simplification. Our approach works in an unsupervised manner that  does not require a parallel corpus for training. In future work, we plan to add paraphrase generation to generate diverse simple sentences.

\section*{Acknowledgments}

We acknowledge the support of the Natural Sciences and Engineering Research Council of Canada (NSERC), under grant Nos.~RGPIN-2019-04897, RGPIN-2020-04465, and the Canada Research Chair Program. 
Lili Mou is also supported by AltaML, the Amii
Fellow Program, and the Canadian CIFAR AI Chair Program. This research was supported in part by Compute Canada (\url{www.computecanada.ca}).

\bibliography{acl2020}
\bibliographystyle{acl_natbib}

\end{document}